\documentclass[10pt,twocolumn,letterpaper]{article}

\usepackage{iccv}
\usepackage{times}
\usepackage{epsfig}
\usepackage{graphicx}
\usepackage{amsmath}
\usepackage{amssymb}

\usepackage{relsize}
\usepackage{url}
\usepackage{tabularx, booktabs}
\usepackage{caption}
\usepackage{pifont}
\usepackage{tikz}
\usepackage{amsmath}
\usepackage{amsthm}
\usepackage{multirow}
\definecolor{lightgray}{gray}{0.9}
\definecolor{lightblue}{rgb}{0.93,0.95,1.0}
\definecolor{darkgreen}{rgb}{0.0,0.6,0.0}
\definecolor{mypink1}{rgb}{0.858, 0.188, 0.478}

\newcolumntype{x}[1]{>{\centering\arraybackslash}p{#1pt}}

\newcommand{\app}{\raise.17ex\hbox{$\scriptstyle\sim$}}

\newlength\savewidth\newcommand\shline{\noalign{\global\savewidth\arrayrulewidth
  \global\arrayrulewidth 1pt}\hline\noalign{\global\arrayrulewidth\savewidth}}
\newcommand{\tablestyle}[2]{\setlength{\tabcolsep}{#1}\renewcommand{\arraystretch}{#2}\centering\footnotesize}

\usepackage[pagebackref=true,breaklinks=true,letterpaper=true,colorlinks,bookmarks=false]{hyperref}

 \iccvfinalcopy % *** Uncomment this line for the final submission

\def\iccvPaperID{7097} % *** Enter the ICCV Paper ID here

\ificcvfinal\fi

\begin{document}
%%%%%%%%% TITLE
\title{ DLT: Conditioned layout generation with \\Joint Discrete-Continuous Diffusion Layout Transformer}

\author{
Elad Levi, \,\,
Eli Brosh, \,\,
Mykola Mykhailych, \,\,
Meir Perez \vspace{3pt}\\
Wix.com\\
% {\tt\small $^1$\{roeiherzig, gamir\}@mail.tau.ac.il}
% \quad \tt\small $^2$\{elad, elibrosh\}@getnexar.com \\
% \quad \tt\small $^3$\{huijuan, trevor\}@cs.berkeley.edu \\
% \tt\normalsize \href{https://embodiedqa.org}{embodiedqa.org}
}

% \and

\maketitle
\ificcvfinal\thispagestyle{empty}\fi

%%%%%%%%% ABSTRACT
\begin{abstract}

Generating visual layouts is an essential ingredient of graphic design. The ability to condition layout generation on a partial subset of component attributes is critical to real-world applications that involve user interaction.
%the inspiration. 
% Diffusion models do not 
Recently, diffusion models have demonstrated high-quality generative performances in various domains. However, it is unclear how to apply diffusion models to the natural representation of layouts which consists of a mix of discrete (class) and continuous (location, size) attributes.
%the idea
To address the conditioning layout generation problem, we introduce DLT, a joint discrete-continuous diffusion model. DLT is a transformer-based model which has a flexible conditioning mechanism that allows for conditioning on any given subset of all the layout component classes, locations, and sizes. %the befit and application (rewrite)
Our method outperforms state-of-the-art generative models on various layout generation datasets with respect to different metrics and conditioning settings. Additionally, we validate the effectiveness of our proposed conditioning mechanism and the joint continuous-diffusion process. This joint process can be incorporated into a wide range of mixed discrete-continuous generative tasks.
\end{abstract}
%%%%%%%%% BODY TEXT
\section{Introduction}
An essential aspect of graphic design is layout creation: the arrangement and sizing of visual components on a canvas or document. A well-designed layout enables users to easily comprehend and efficiently interact with the information presented. The ability to generate high-quality layouts is crucial for a range of applications, including user interfaces for mobile apps and websites~\cite{RICO} and graphic design for information slides \cite{presentationSlides}, magazines \cite{MAGAZINE}, scientific papers \cite{PubLayNet}, infographics \cite{infographics}, and indoor scenes \cite{indoorScenes}.

\begin{figure}[t!]
	\begin{center}
        \includegraphics[width=0.9\linewidth]{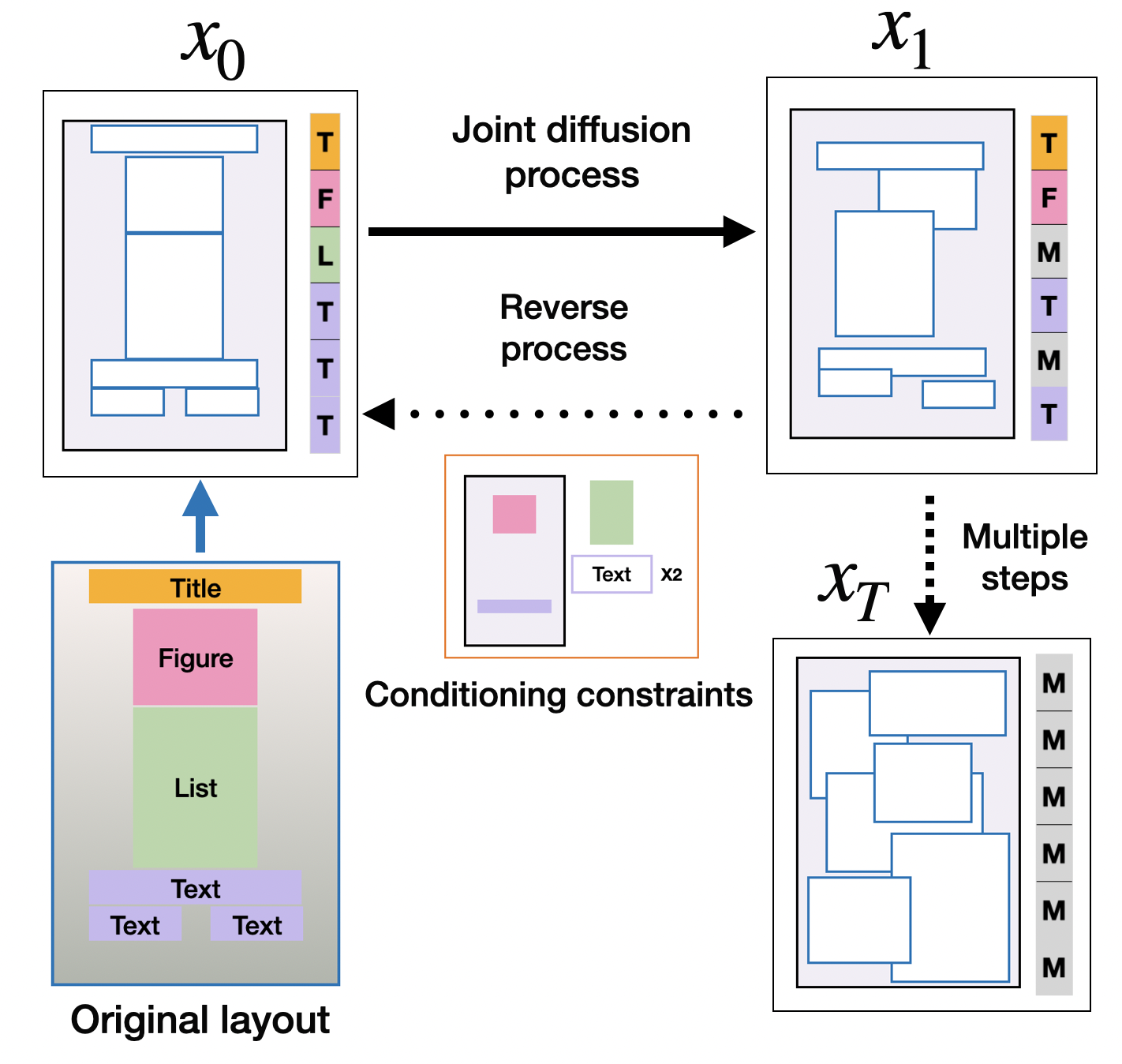}
    %\vspace{-10pt}
	\caption{\small{{Overview of our Joint Discrete-Continuous diffusion process for layout generation. The diffusion process is applied jointly on both the continuous attributes of the components (size and location) by adding a small Gaussian noise, and on the discrete components attribute (class) by adding a noise that adds a mass to the Mask class. The reverse diffusion model can be conditioned on any given subset of the attributes.}}}
	\label{fig:overview}
	\end{center}
\end{figure}

To facilitate graphic design tasks, modelling approaches such as Generative Adversarial Networks (GANs)~\cite{LayoutGAN}, Variational Autoencoders (VAEs) \cite{LayoutVAE,VTN}, and masking completion \cite{LayoutBERT,BLT,LT} were explored for generating novel layouts. These approaches employ various architectures such as Recurrent Neural Networks (RNNs)~\cite{LayoutVAE}, and Graph Neural Networks (GNNs)~\cite{LeeJELG0Y20}. Recently proposed, Transformer-based approaches \cite{VTN,LayoutBERT,BLT,LT} have been shown to produce diverse and plausible layouts for a variety of applications and user requirements. 
 
Layouts are often represented as a set of components, each consisting of several attributes, such as class, position, and size~\cite{LayoutGAN}. Inspired by recent advancements in NLP, a common practice is to model a component as a sequence of discrete  attribute tokens, and a layout as the concatenation of all attribute tokens~\cite{LayoutBERT,LT,BLT}. The model is trained using a self-supervised objective (either next-word completion or unmasking) to generate an output sequence. The discretization process is done by binning geometry attributes, such as position and size.
% A common practice in these approaches is to discretize geometry attributes, such as position and size, by quantization or binning. 
However, using discrete tokens fails to take into account the natural representation of (geometry) data, which is continuous, and either result in poor resolution of layout composition, or in a large sparse vocabulary that is not covered well by smaller datasets.

Providing controllable layout generation is crucial for many real-world graphic design applications that require user interaction. We would like to allow a user to fix certain component attributes, such as component class, or a set of components, and then generate the remaining attributes (position and size) or components. In recent iterative-based masking completion approaches~\cite{BLT}, such conditioning is performed only during inference, in a way that does not provide the model with the ability to separate between the conditioned and generated parts of the layout during the iterative refinement. This lack of separation might lead to ambiguity in the generation process which can result in suboptimal performance.

%An important aspect of layout generation is the amount of control offered to the user. We would like to allow a user to fix certain component attributes, such as component class, or a set of components, and then generate the remaining attributes (position and size) or components.  In recent iterative-based approaches~\cite{BLT}, such conditioning is performed only during inference, in a way that does not provide the model with the ability to separate between the conditioned and generated parts of the layout during the iterative refinement. A more effective approach is to train the model to perform layout conditioning and inform the model of the condition/generation separation during the iterative refinement. Inspired by recent diffusion-based image editing techniques~\cite{GLIDE}, we propose a flexible editing mechanism which allows explicit conditioning on any subset of the layout data.  We later show that this design choice improves model performance significantly compared to inference-based conditioning. 
  
Diffusion models are a generative approach that has gained significant attention recently. Continuous diffusion models have demonstrated remarkable generative capabilities in various tasks and domains, such as text-to-image~\cite{DALLE}, video generation~\cite{diffusionVideo}, and audio~\cite{Diffsound}.  While diffusion models were extended to discrete spaces~\cite{DISCE,DISCN}, a challenge still remains on how to apply these models to generative tasks whose representation consists of both continuous and discrete features. 
%In fact, \cite{BLT} which extends layout transformers to enable controllability by implementing a hierarchical masking policy, can be viewed as an approximated Discrete Diffusion process.
In this paper, we introduce a Diffusion Layout Transformer (DLT) for layout generation and flexible editing.  Different from traditional diffusion models, we propose a novel framework based on a joint continuous-discrete diffusion process and provide theoretical justification for the derived optimization objective. Using a transformer-encoder architecture, we apply a diffusion process jointly both on the continuous attributes of the layout components (size and location) and on the discrete ones (class), as depicted in Figure~\ref{fig:overview}.  Our diffusion model addresses an important limitation of transformer-based models that rely on a discrete representation of the layout, which can lead to reduced resolution and limited diversity of layout compositions. 
% Our novelty lies in the proposed mixed discrete-continuous embedding representation and diffusion model, which as demonstrated by our comprehensive evaluation, is essential for generating high-quality layouts. 

% 
 Our framework offers flexible layout editing which allows for conditioning on any subset of the attributes of the layout components  provided (class, size and location). Unlike masking completion approaches~\cite{LayoutBERT,BLT,LT}, which perform conditioning only during inference, and inspired by diffusion image inpainting techniques~\cite{GLIDE}, we explicitly train the model to perform layout conditioning. We use condition embedding to control on which layout attributes to apply the diffusion process. In this way, the model is able to separate between the condition and the generation parts during the iterative refinement.
 Our experiments reveal that these design choices significantly improve  the model's performance compared to inference-based conditioning.

We study the effectiveness of our DLT model by evaluating it on three popular layout datasets of diverse graphic design tasks over several common metrics.  Using extensive experiments, we demonstrate that our proposed model outperforms state-of-the-art layout generation models on layout synthesis and editing tasks while maintaining runtime complexity on par with these models. By comparing different alternatives, we show that both the joint diffusion process and the conditioning mechanism contribute to the model's strong generative capabilities. Our joint discrete-continuous diffusion framework has a generic design and thus is applicable to other domains such as multi-modal generation tasks (text+image, text + audio, etc'). 

%To summarize, our main contributions are:
%\begin{enumerate}
%    \item We propose a novel method for high-quality layout generation based on a joint continuous-discrete diffusion framework with powerful layout editing.
%    \item We conduct an extensive evaluation  on diverse layout benchmarks and demonstrate that our method outperforms state-of-the-art models on commonly-used quality metrics.
%\end{enumerate}

\section{Related work}
\textbf{Layout synthesis.} The task of generating high-quality, realistic and diverse layouts using data-driven generative methods has received increased attention in %increased, considerable? \myk{\cite{LeeJELG0Y20,Docsim,infographics,presentationSlides,LayoutGAN,LayoutVAE,LayoutBERT,LT,VTN,BLT}}
recent years. Several generative approaches have been proposed to address this problem.  For example, LayoutGAN~\cite{LayoutGAN} trains a generator to map a gaussian/uniform distribution to the locations/categories of layout components by rendering a wireframe and training a pixel-based discriminator. LayoutVAE~\cite{LayoutVAE} introduces two types of Variational Auto-Encoders. The first one learns the distribution of category counts of components, while the second one predicts the locations and sizes of components, conditioning on the categories. Motivated by the recent advancements in NLP, transformer-based language models were proposed to solve the layout generation task. In~\cite{LT}, a standard transformer-decoder model was applied to generate the attributes of layout components, where these components are sorted in lexicographic order. Training is performed with a self-supervised objective of predicting the next attribute given the attributes of previous components. In BLT~\cite{BLT}, a bidirectional transformer encoder was used to predict the attributes of layout components. In order to improve the quality of the generated layouts, they iteratively refine  layout attributes by remasking the attributes with the lowest probability. Unlike previous methods, our generative method is based on a diffusion process (see Section~\ref{sec:model}), which has demonstrated strong generative capabilities in various fields.

\textbf{Layout conditioning.} In order to be applicable to many real-world graphic design applications, where user interaction is required, various methods were proposed for controllable layout generation. LayoutGAN++~\cite{LayoutConstrianed} improves LayoutGAN~\cite{LayoutGAN} by replacing the generator and the discriminator with a transformer-based architecture. They propose to add constraints on the layout generation by defining and solving a nonlinear optimization problem in the latent space. In the case of VAE, Lee et al~\cite{LeeJELG0Y20} proposed a method to constraint the layout generation on a set of relative design properties using a graph-based conditional VAE model. VTN~\cite{VTN} is a transformer-based conditional VAE approach in which the conditioning is performed either by an RNN that maps the condition to a prior distribution or by a transformer encoder. Different from these conditioning approaches which impose restrictions in the latent space, in BLT~\cite{BLT} the conditioning is  done simply by unmasking the conditioned component attributes. Similar to~\cite{BLT}, our approach also allows the same flexible controllable conditioning, and during inference, the model is explicitly informed of the conditioning attributes. However, we propose to perform the optimization of the conditioning scenarios during training, which enhances the model's performance.

\textbf{Diffusion Generative Models.}
Diffusion models are a family of neural generative models which parameterize a reverse Markov noising process~\cite{DEF_orig}. In the continuous diffusion case, a small amount of Gaussian noise is added to the sample. Continuous diffusion models have shown strong generative capabilities across diverse tasks and modalities such as text-to-image~\cite{GLIDE,DALLE,IMAGEN,StableDiffusion}, video generation~\cite{MakeVid,ImagenVideo}, text-to-3D~\cite{DreamFusion,Magic3D} and audio~\cite{Diffsound,AudioLDM,MakeAudio}. Recently, transformer-based diffusion models have demonstrated state-of-the-art performance in various tasks such as text-to-image generation~\cite{transformerdiff} and human motion generation~\cite{MDM}. Inspired by these works, our diffusion model is also based on a transformer-encoder architecture.

\textbf{Discrete diffusion process.} In order to generalize the diffusion process to discrete cases, Austin et al.~\cite{DISCN} suggest applying the Markov noising process on the probability vector of categories. The effectiveness of this method was demonstrated in~\cite{DISCE} on the image captioning task, where the stationary distribution has all the mass on the [MASK] absorbing class. The natural representation of the layout generation task contains both continuous (size and location) and discrete (category) component attributes. Different from previous approaches which apply either continuous or discrete diffusion process, our approach applies the diffusion process on the joint distribution. 

\section{Method}

\begin{figure*}[t!]
	\begin{center}
        \includegraphics[width=0.99\linewidth]{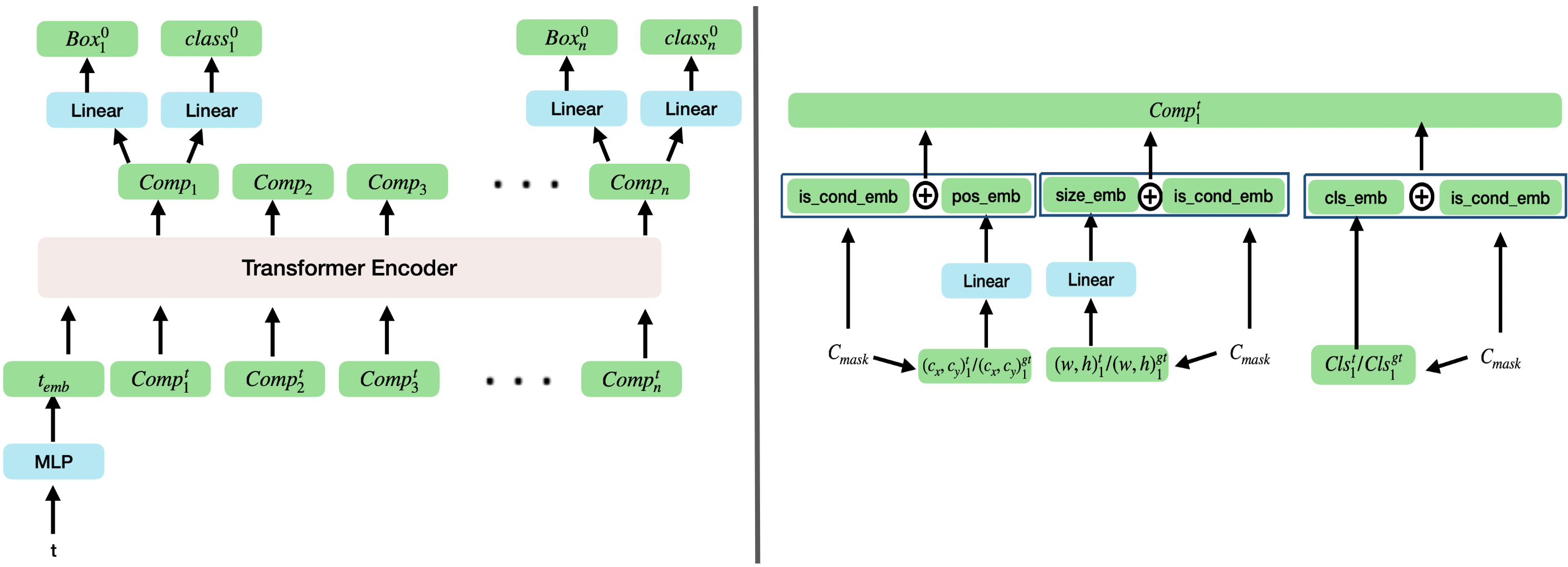}
    \vspace{-10pt}
	\caption{\small{ DLT model architecture \textbf{Left:} The transformer is fed with an embedding that represents step $t$ of the diffusion process and an embedding for each of the layout components. The output is the cleaned  coordinates and classes of components. During inference, the output is noised back to step $t-1$.
	\textbf{Right:} The component embedding consists of a concatenation of position embedding, size embedding and class embedding. Each one of the three sub-embeddings can be either generated from the diffusion process or from the conditioning information. The model is explicitly informed which part of the input is part of the diffusion process by adding a trainable vector to the sub-embedding in case the attribute is part of the conditioning. This condition embedding is separate for each attribute. 
	%This allows to refine the even/odd classes to their intra-class digits.
	}}
	\vspace{+10pt}
	\label{fig:arch}
	\end{center}
\end{figure*}

Our goal is to generate layouts conditioning on constraints $c$. The layout is represented by a set of $N$ components $\{B_i\}^N_{i=1}$, where each component $B_i$ consists of a bounding box describing the location and size of the component in the layout, and the component class.  Similar to~\cite{BLT}, the condition is a subset of all locations, sizes and classes of the layout components. In particular, unconditional layout generation is also supported by setting $c=\emptyset$.
\subsection{Diffusion model}
Following the framework in~\cite{DEF_orig}, a diffusion process is modeled as a Markov noising process with $T$ steps $\{\bar{x}_t\}^T_{t=0}$, where $\bar{x}_0\sim q(\bar{x}_0)$ is sampled from the real data distribution, and the forward process is sampled according to $\bar{x}_{t+1} \sim q(\bar{x}_{t+1} | \bar{x}_t)$, where $q(\bar{x}_{t+1} | \bar{x}_t)$ is a predefined (and known) distribution. A diffusion model defines the parameterized reverse diffusion process as: \[p_\theta(\bar{x}_{0:T}) = p(\bar{x}_T)\prod_{t=1}^Tp_\theta(\bar{x}_{t-1}| \bar{x}_{t})\]
Fitting $p_{\theta}(\bar{x}_0)$ to match the real data distribution $q(x_0)$ is done by optimizing a variational bound on the negative log-likelihood. This variational bound can be expressed as a sum of three terms (see eq.(5) in~\cite{DEF_orig}): \[\begin{cases}
L_T = D_{KL}(q(\bar{x}_T | \bar{x}_0)|| p(\bar{x}_T))\\
L_{t} = D_{KL}(q(\bar{x}_{t-1} | \bar{x}_t,\bar{x}_0)|| p_\theta(\bar{x}_{t-1} | \bar{x}_t)) & 1\leq t \leq T-1 \\ 
L_0 = -log(p_\theta(\bar{x}_{0} | \bar{x}_{1}))
\end{cases}\]
Since $L_T$ is constant it can be ignored in the optimization process.

\textbf{Continuous diffusion models.} We denote the functions in the continuous process with the superscript $c$. Following~\cite{DEF_orig}, the forward process is performed by adding a Gaussian noise \[q^{c}(\bar{x}_t | \bar{x}_{t-1}) =  \mathcal{N}(\bar{x}_t, \sqrt{1-\beta_t} \bar{x}_{t-1}, \beta_t\cdot I), \]
where a constant variance $\beta_t\in (0,1)$ (a hyper-parameter) controls the sizes of the steps. When $T\rightarrow \infty$ we can approximate $\bar{x}_T \sim \mathcal{N}(0,I)$. In this case, $p^c_\theta(\bar{x}_{t-1} | \bar{x}_{t})$ is parameterized by a family of Gaussian distributions, and $L^{c}_{t}$ can be computed in a closed form. 

In our setting, we apply the diffusion process on a set of all the bounding boxes in the layout: $\bar{x}_0 \in \mathbb{R}^{NX4}$, where $N$ is the maximal number of components. For the reverse diffusion process, we follow~\cite{DALLE,MDM}, and train a model $F^c_\theta(\bar{x}_t,c,\bar{y}_t)$ to directly predict $\bar{x}_0$ from the noise, where $\bar{y}_t$ carries the class information in the diffusion process, as described in the next paragraph. The final loss $\mathcal{L}_{box}$ is a reweighted version of the loss $L^{c}_t$, which simplifies the objective and empirically leads to better results:
\[\mathcal{L}_{box} = \mathbb{E}_{\bar{x}_0,\bar{y}_0 \sim q(\bar{x}_0 , \bar{y}_0| c), t\sim[0,1]} ||F^c_\theta (\bar{x}_t,c,\bar{y}_t)- \bar{x}_0||^2 \] 
\textbf{Discrete diffusion models.} We denote the functions in the discrete process with the superscript $d$.
The discrete diffusion process was defined in~\cite{DISCN}. In this setting, we have discrete random variables  with K categories, $y_0 \ldots,y_T \in {\{1,\ldots,K\}}$. $y_0$ is drawn from the data distribution, and the forward noising process is defined by the Markov transition matrix: \[ [Q_t]_{i,j} = q^d(y_t = i | y_{t-1} = j) \]
In our case, we apply the discrete diffusion process on the set of all the classes of layout components $\bar{y}_0 \in \{0,..,K\}^N$. We follow~\cite{DISCN, DISCE} and choose a transition matrix with an absorbing state as the K-th category, denote by [Mask]. In this case for $i<K$:
% \[ [Q_t]_{i,j} = \begin{cases}
% \alpha & i=j\\
% \beta & i=K \: ([Mask]) \\
% \frac{1-(\beta - \alpha)}{K-2}  & else
% \end{cases} \]
% and for $i=K$:
% \[ [Q_t]_{K,j} = \begin{cases}
% 1 & j=K\\
% 0 & j\neq K
% \end{cases} \]

\[ [Q_t]_{i,j} = \begin{cases}
\beta & i=j\\
1-\beta & i=K \: ([Mask]) 
\end{cases} \]
and for $i=K$:
\[ [Q_t]_{K,j} = \begin{cases}
1 & j=K\\
0 & j\neq K
\end{cases} \]

In other words, if the class is [Mask], it will remain the same. If not, then with probability $\beta$ the class remains the same, and with probability $1-\beta$ it will be changed to $[Mask]$. Observe that when $T\rightarrow \infty$ we can approximate $\bar{y}_T = [Mask]^N$.    
In the reverse diffusion process, we train the model  $F^d_\theta(\bar{x}_t,c,\bar{y}_t)$ to directly predict the class probabilities of the components $\Tilde{P_\theta}(\Tilde{y_0} | \bar{y}_t,\bar{x}_t,c)$. The parameterization of the reverse diffusion probability is defined using the forward process \[p^d_\theta (\bar{y}_{t-1}|\bar{y}_{t},\bar{x}_t,c) =  \sum_{\Tilde{y}_0\in K^N} q(\bar{y}_{t-1}| \bar{y}_{t},\Tilde{y}_0)\cdot \Tilde{P_\theta}(\Tilde{y_0} | \bar{y}_t,\bar{x}_t,c)\]
In this case, the regular cross-entropy loss is a reweighted function of $L^{d}_t$ (see A.3 in~\cite{DISCN}):
\[\mathcal{L}_t = CE(F^d_\theta (\bar{x}_t,c,\bar{y}_t), \bar{y}_0)\] Following \cite{DISCE}, our objective on the class category is
\[\mathcal{L}_{cls} = \mathbb{E}_{\bar{y}_0,\bar{x}_0 \sim q(\bar{y}_0,\bar{x}_0| c), t\sim[0,1]} CE(F^d_\theta (\bar{x}_t,c,\bar{y}_t), \bar{y}_0) \]

\begin{table*}[h!]
\centering

\tablestyle{5.2pt}{1.25}
\begin{tabular}{lx{22}x{22}x{22}x{22}|x{22}x{22}x{22}x{22}|x{22}x{22}x{22}x{22}}
\shline
\multicolumn{1}{c}{Dataset} & \multicolumn{12}{c}{ \textbf{Publaynet}} \\\cline{2-13}

\multicolumn{1}{c}{} & \multicolumn{4}{c}{Conditioned on Category} & \multicolumn{4}{c}{Category + Size}& \multicolumn{4}{c}{Uncoditioned} \\ Model
& pIOU & Overlap & Alignment & FID & pIOU & Overlap & Alignment & FID &  pIOU & Overlap & Alignment & FID \\
\shline
LT~\cite{LT} & 2.7 & 7.6 & 0.41 & 26.8 & 7.1 & 11.7 & 0.14 & 22.0 & 0.62 & 2.4 & 0.11 & 19.3 \\
BLT~\cite{BLT} & 0.89 & 4.4 & \textbf{0.10} & 36.6 & 1.7 & 8.1 & \textbf{0.09} & 14.2 & 0.60 & 2.7 & 0.12 & 69.8\\
VTN~\cite{VTN}& 2.1 & 6.8 & 0.29 & 22.1 & 5.3 & 15.3 & \textbf{0.09} & 17.9 & 0.68 & \textbf{2.6} & \textbf{0.08} & 14.5 \\
\cline{1-13}
DLT  & \textbf{0.67} & \textbf{3.8} & 0.11 & \textbf{10.3} & \textbf{0.82} & \textbf{4.2} & \textbf{0.09} & \textbf{11.4} & \textbf{0.59} & \textbf{2.6} & 0.11 & \textbf{13.8}\\
\shline

\multicolumn{1}{c}{Dataset} & \multicolumn{12}{c}{ \textbf{Rico}} \\\cline{2-13}

\multicolumn{1}{c}{} & \multicolumn{4}{c}{Conditioned on Category} & \multicolumn{4}{c}{Category + Size}& \multicolumn{4}{c}{Uncoditioned} \\ Model
& pIOU & Overlap & Alignment & FID & pIOU & Overlap & Alignment & FID &  pIOU & Overlap & Alignment & FID \\
\shline
LT~\cite{LT} & 25.6 & 75.2 & 0.58 & 14.7 & 23.8 & \textbf{69.1} & 0.41 & 8.4 & 23.2 & 65.1 & 0.40 & 15.2 \\
BLT~\cite{BLT} & 30.2 & 85.1 & \textbf{0.12} & 27.8 & 24.5 & 79.3 & 0.30 & 10.2 & 23.0 & 70.6 & 0.25 & 18.7\\
VTN~\cite{VTN}& 25.4 & 74.2 & 0.43 & 14.3 & 24.1 & 69.6 & 0.44 & 7.1 & 29.4 & 72.1 & 0.26 & 29.4\\
\cline{1-13}
DLT  & \textbf{21.9} & \textbf{70.6} & 0.18 & \textbf{9.5} & \textbf{17.2} & 70.2 & \textbf{0.28} & \textbf{6.3} & \textbf{19.3} & \textbf{58.4} & \textbf{0.21} & \textbf{13.9} \\
\shline

\multicolumn{1}{c}{Dataset} & \multicolumn{12}{c}{ \textbf{Magazine}} \\\cline{2-13}

\multicolumn{1}{c}{} & \multicolumn{4}{c}{Conditioned on Category} & \multicolumn{4}{c}{Category + Size}& \multicolumn{4}{c}{Uncoditioned} \\ Model
& pIOU & Overlap & Alignment & FID & pIOU & Overlap & Alignment & FID &  pIOU & Overlap & Alignment & FID \\
\shline
LT~\cite{LT} & 19.9 & 71.0 & 1.5 & 44.7 & 21.4 & 70.2 & \textbf{1.2} & 45.3 & 21.4 & 70.0 & 1.1 & 42.6 \\
BLT~\cite{BLT} & 36.4 & 133 & 1.4 &49 & 20.5 & 56.8 & \textbf{1.2} & 27.3 & 30.1 & 134 & 1.1 & 52.7\\
VTN~\cite{VTN} & 10.3& 38.7 & 2.4 & 37.6 & 9.9 & 28.8 & 2.3 & 29.4 & 20.1 & 70.7 & \textbf{0.9} & 62.7\\
\cline{1-13}
DLT  &  \textbf{5.9} & \textbf{16.1} & \textbf{1.3} & \textbf{26.2} & \textbf{6.8} & \textbf{19.4} & 1.6 & \textbf{21.7} & \textbf{4.8} & \textbf{12.1} & 1.8 & \textbf{40.9}\\
\shline

\end{tabular}

\vspace{+10pt}
\caption{Quantitative comparison of layout generation with various constraints setting. The values of Alignment, Overlap, and pIOU are multiplied by 100 for clarity. The DLT method achieve overall the best results among all tested model.}
\label{table:1}
\end{table*}

\textbf{Combined diffusion process.} Our layout diffusion process is obtained by sampling $(\bar{x}_0,\bar{y}_0)$ from the data distribution. The forward diffusion process is obtained by applying the continuous process $q^c$ and the discrete process  $q^d$ independently on $\bar{x}_0, \bar{y}_0$. The model is trained with the combined objective:
\[\mathcal{L}_{model} = \lambda_1\cdot \mathcal{L}_{box} + \lambda_2\cdot \mathcal{L}_{cls} \]
The theoretical motivation for the combined loss is explained by parameterizing the reverse process joint distribution as independent over the previous step: 
\[p_\theta(\bar{x}_{t-1}, \bar{y}_{t-1} |\bar{x}_{t}, \bar{y}_{t}) = p^c_\theta(\bar{x}_{t-1}, |\bar{x}_{t}, \bar{y}_{t})\cdot  p^d_\theta(\bar{y}_{t-1}, |\bar{x}_{t}, \bar{y}_{t})   \]
% This combined loss can be explained theoretically by assuming that $\bar{x}_{t-1}$ is independent from $\bar{y}_{t-1}$ over $\{c, \bar{x}_{t},\bar{y}_{t}\}$. 
Observe that in this decomposition, the probabilities of the continuous part and the discrete part are still dependent on both the continuous coordinates and the discrete classes of the previous diffusion step.
With this parametrization, by the additive property of the KL-divergence for independent distributions we have:
\[L_t = L^{c}_t + L^{d}_t\]
Therefore, our final loss is a simplified reweight optimization of the variational bound of the joint distribution negative log-likelihood function.

% \begin{align*}
%   L_t = D_{KL}(q(\bar{x}_{t-1},\bar{y}_{t-1} | \bar{x}_t,\bar{y}_{t},\bar{x}_0),\bar{y}_{0}||\\
%   =\\
%   kjkj
% \end{align*}

\textbf{Sampling.} In the inference step, we sample $\bar{x}_0 \sim \mathcal{N}(0,I)$, $\bar{y}_0 = [Mask]^N$. In each step $t$, we apply $F^c_\theta(\bar{x}_t,c,\bar{y}_t)$, $F^d_\theta(\bar{x}_t,c,\bar{y}_t)$, getting $\bar{x}_{pred}$, $\bar{y}_{pred}$ (where we keep class prediction as probability vector). We then apply the noise and sample according to 
\[(\bar{x}_{t-1}, \bar{y}_{t-1})\sim p_\theta(\bar{x}_{t-1}, \bar{y}_{t-1}| \bar{x}_{t}, \bar{y}_{t},c)\]
\subsection{Model} \label{sec:model}
Our DLT model is illustrated in  Figure~\ref{fig:arch}. Similarly to~\cite{MDM} the backbone of $F_\theta$ is a multi-layer Transformer encoder~\cite{Attention_all_you_need}. The input for the transformer network is a set of embedding vectors of the components and an embedding that encodes the process's time-step $t$. This time embedding is a result of an MLP network that injects the time into the transformer embedding dimension. The transformer encoder outputs an embedding for each component, which is fed to the box head and the class head, both of which are a simple Linear layer. The box head predicts directly the box coordinates, while the class head predicts the class logits.

The class embedding consists of a concatenation of three different embeddings: (1) The position embedding, (2) the size embedding, and (3) the class embedding. The position and size embeddings are implemented by a Linear layer that takes the center and size of the components, respectively. The class embedding is implemented as an embedding layer that holds a separate embedding vector for each class.
\subsection{Conditioning Mechanisms} \label{sec:conditioning} 

% \begin{figure}[t!]
% 	\begin{center}
%         \includegraphics[width=1\linewidth]{iccv2023AuthorKit/figures/condition_embedding2.png}
%     %\vspace{-10pt}
% 	\caption{\small{\textbf{Component embedding structure.} The component embedding consists of a concatenation of position embedding, size embedding and class embedding. Each one of the three sub-embeddings can be either generated from the diffusion process or from the conditioning information. The model is informed explicitly which part of the input is part of the diffusion process by adding a trainable vector to the sub-embedding.}}
% 	\label{fig:conditioning}
% 	\end{center}
% \end{figure}

We propose to perform layout editing using a flexible conditioning architecture, that allows the user to choose the specific layout information that is given as part of the conditioning, a capability similar to that in~\cite{BLT}. Following early diffusion image-inpainting works~\cite{Inpainting_b1, Inpainting_b2, Inpainting_b3}, the editing process in~\cite{BLT} is performed only in the inference phase by replacing the known information with the noise in the reverse-diffusion process. In contrast, inspired by the recent advancement in the diffusion image inpainting field~\cite{GLIDE}, we explicitly train the model to perform the layout editing. We perform editing by adding a condition embedding (implemented as a binary embedding layer), that informs the model which part of the input is conditioned and which one is part of the diffusion process. This architecture is illustrated in Figure~\ref{fig:arch}. During training, we condition on a random subset of all components' locations, sizes, and classes. We apply the loss only on the unconditioned information.  
\section{Experiments}
In this section, we demonstrate the effectiveness of our DLT model, by evaluating our proposed method on 3 diverse layout benchmarks. Our proposed model is able to outperform state-of-the-art layout generation models on layout synthesis and editing tasks with respect to various metrics that examine realism, alignment, and overlapping.
\subsection{Datasets}
We evaluate our model on three popular public layout generation datasets that represent different graphic design tasks. \textit{RICO}~\cite{RICO,RICO2} provides user interface designs for mobile applications. It contains 91K layouts with 27 different component types. \textit{PubLayNet}~\cite{PubLayNet} contains 330K samples of scientific documents. The layout components consist of five categories: text, title,
figure, list, and table. \textit{Magazine}~\cite{MAGAZINE} contains 4K magazine pages with components from six categories (texts, images, headlines, over-image texts, over-image headlines,
and backgrounds). We follow the evaluation protocol of previous works~\cite{LeeJELG0Y20,LayoutGAN,LayoutConstrianed}, and exclude layouts with non-top 13 category components in the RICO dataset. We also exclude layouts with more than 10 components in both the RICO and PubLayNet datasets. Following \cite{LayoutConstrianed}, we use 85\% of the RICO dataset for training, 5\% for validation, and 10\% for testing. In PubLayNet and Magazine, we used the official data split.

\begin{table*}[!h]
\centering

\tablestyle{5.2pt}{1.1}
\begin{tabular}{lx{33}x{33}x{33}|x{33}x{33}x{33}}
\shline
\multicolumn{1}{c}{\textbf{Docsim metric}} & \multicolumn{3}{c}{Conditined on Category} & \multicolumn{3}{c}{Category + Size} \\ 
  Model & Publaynet & RICO & Magazine & Publaynet & RICO & Magazine \\
\shline
LT~\cite{LT} & 9.3\scriptsize{$\pm 0.17$}&15.9\scriptsize{$\pm 0.52$} &10.8\scriptsize{$\pm 0.62$} &18.5\scriptsize{$\pm 0.19$}&26.1\scriptsize{$\pm 0.71$}&15.5\scriptsize{$\pm 0.79$}\\
BLT~\cite{BLT} &  14.4\scriptsize{$\pm 0.21$}& 16.4\scriptsize{$\pm 0.21$} &12.4\scriptsize{$\pm 0.70$} &20.7\scriptsize{$\pm 0.13$}&25.8\scriptsize{$\pm 0.56$}&19.5\scriptsize{$\pm 0.95$}\\
VTN~\cite{VTN} & 9.8\scriptsize{$\pm 0.23$} &18.5\scriptsize{$\pm 0.46$} & 8.5\scriptsize{$\pm 0.65$}&19.7\scriptsize{$\pm 0.21$} & 28.1\scriptsize{$\pm 0.67$} & 7.8\scriptsize{$\pm 0.83$} \\
\cline{1-7}
DLT  & \textbf{16.2}\scriptsize{$\pm 0.23$}&\textbf{20.4}\scriptsize{$\pm 0.42$} &\textbf{13.5}\scriptsize{$\pm 0.72$} &\textbf{24.3}\scriptsize{$\pm 0.17$}&\textbf{28.8}\scriptsize{$\pm 0.48$}&\textbf{22.7}\scriptsize{$\pm 0.81$}\\
\shline

\end{tabular}

% \vspace{-10pt}
\caption{Docsim metric results on the tested datasets in the two conditioning scenarios, results are multiplied by 100 for clarity. Our method outperforms all tested models both when conditioning on category and when conditioning on category+size.}
\label{table:2}
\end{table*}

\subsection{Evaluation metrics}
We use five common metrics in the literature to evaluate the quality of the generated layout: \textit{Fréchet inception distance} (FID) \cite{FID}, which  measures the distance between the generated layout distribution and the real layout. Following~\cite{LeeJELG0Y20}, we compute the score by training a network to classify between real layouts and layouts with additional injected noise (for this, we used the intermediate features of the network). \textit{Overlap} \cite{LayoutGAN} measures the total overlapping area between any two components. \textit{Perceptual IOU} (pIOU) \cite{BLT} computes the overlap of each component divided by the union area of all objects. \textit{Alignment score}~\cite{LeeJELG0Y20} measure the alignments between pair of components, where the assumption is that in graphic design every component has to be aligned to at least one of the other components either by the center or by one of the sides. We used \textit{DocSim}~\cite{Docsim} to measure the similarity between the generated layouts and the conditioned layouts.

\subsection{Conditioning setting}
Following~\cite{BLT}, we tested three different conditioning layout generation scenarios: \textit{Category} conditioning, where only the component categories are provided and the model has to predict their size and location in the layout; \textit{Category + Size} conditioning, where the model is given also the size of the components and only needs to predict the position; and \textit{Unconditioned} generation, where no information is provided to the model.
\subsection{Compared methods}
We compare our proposed method to state-of-the-art layout generation models. 
LayoutTransformer (LT) \cite{LT} is based on an autoregressive transformer architecture. For the implementation, we used nucleus sampling \cite{Nucleus} with $p=0.9$, which  significantly improves the diversity of the results. In order to support the category + size conditioning, we changed the order of components tokens to $(w,h,x,y)$ instead of $(x,y,w,h)$, and verified that this change does not affect the model's performance.
For the BLT \cite{BLT} implementation, we used top-k sampling (k = 5) only in the unconditioned setting as suggested by~\cite{BLT}. We also compare ourselves  to the Variational Transformer Network (VTN)~\cite{VTN}, a conditional VAE method.

\subsection{Implementation details}
\label{4.5}
The transformer encoder was trained with 4 layers, 8 attention heads and a latent dimension of 512. For the component location, we used the same pre-process as in~\cite{DiffusionDet}. The diffusion model was trained with a cosine noise schedule~\cite{cosine}. For the discrete process, we used the hyper-parameter $\beta=0.15$. We set the number of discrete diffusion steps to $T/10$ (where $T$ is the number of the continuous steps). During training and inference, we synced between the discrete and the continuous process by taking $Q^{t//10}$, where $0\leq t\leq T$ is the continuous step. For the loss hyper-parameters, we used $\lambda_1 = 5$ and $\lambda_2 = 1$.

During training, for each sample, we uniformly sample between the three different conditioning settings: Category, Category + size, and unconditioned. We mask the loss on the outputs which correspond to the conditioned inputs.

\begin{figure}[t!]
    \begin{minipage}[t]{0.8\linewidth}
        \centering
        \scalebox{0.4}{\input{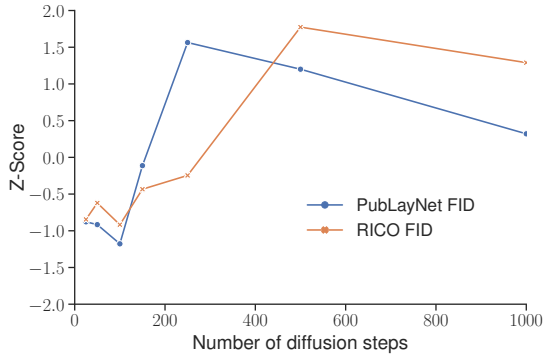}}
       
        % \caption{This is caption for Figure 1.}
    \end{minipage}
    \caption{Z-score of the FID metric across
a varying number of diffusion steps. Best performances were achieved with 100 diffusion steps. }

     \label{fig:steps}
\end{figure}

\subsection{Number of diffusion steps}
The performance of the model over different numbers of diffusion steps can be seen in Figure~\ref{fig:steps}. It is important to note that the DLT inference time is a function of the number of diffusion steps and is not dependent on the number of components, unlike the other models we tested. Given the results in  Figure~\ref{fig:steps}, we choose $T=100$ diffusion steps for all our tested experiments. This means that our model inference time is $\sim2$ times slower on the RICO and Publaynet datasets. However, our model is $\sim1.65$ times faster on the Magazine dataset.

\begin{figure*}[t!]
	\begin{center}
        \includegraphics[width=0.80\linewidth]{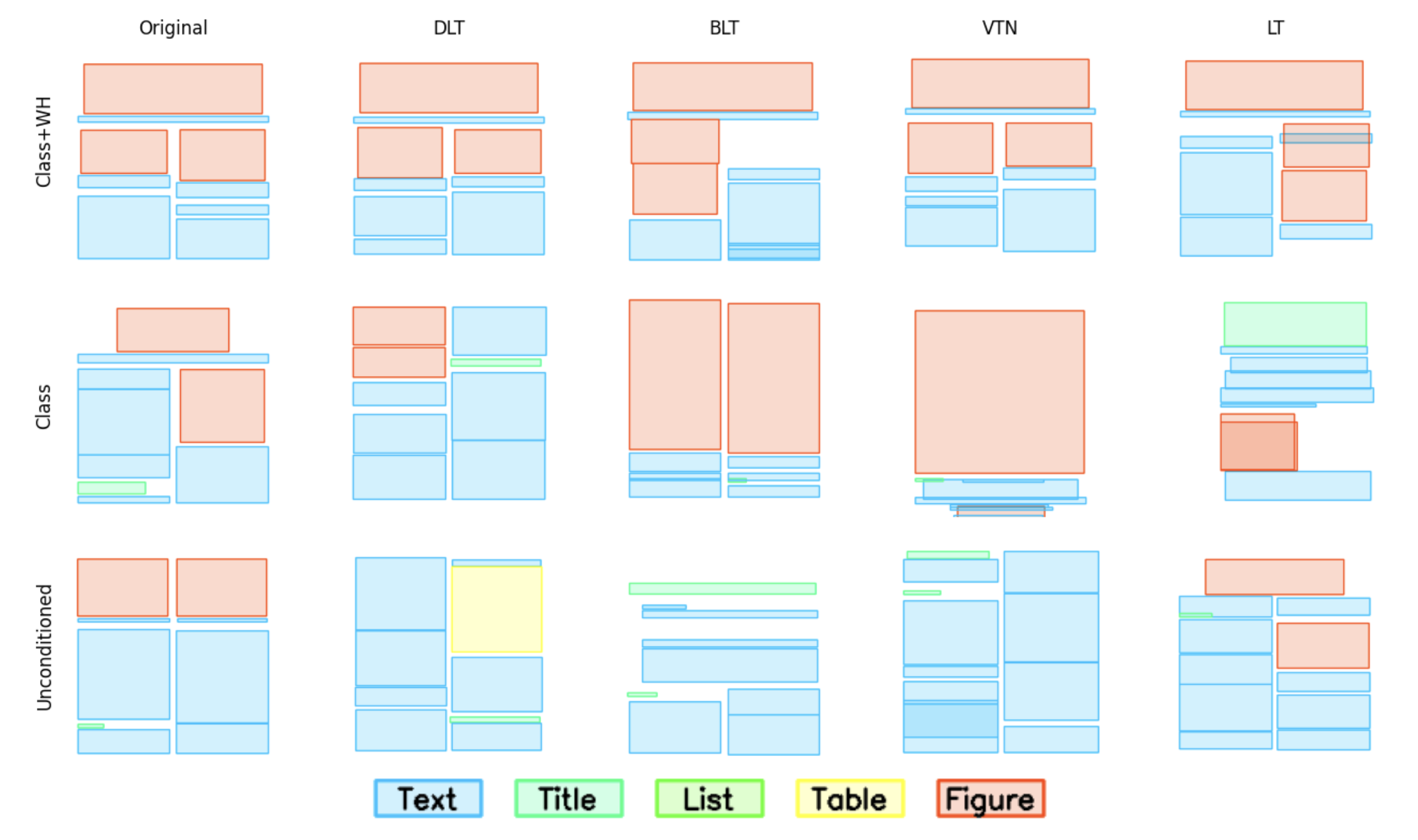}
    \vspace{-10pt}
	\caption{\small{Qualitative comparison of generated layouts between all tested methods on PubLayNet dataset with the three tested conditioning settings. 
	}}
	% \vspace{-10pt}
	\label{fig:res}
	\end{center}
\end{figure*}

\subsection{Results}
Each model was tested with four trials (on the test set). We report the averaged results of the metrics in Table \ref{table:1}, and the standard deviation results in Appendix A. While the BLT model performs better than the LT model in the conditioning scenarios, and the LT model performs better in the unconditioned setting, our model outperforms the tested methods by statistically significant margins in both scenarios with respect to all tested metrics. The only exception is the alignment score where in some of the cases the BLT model performs better. This is due to the small number of bins (32) in the discretization of the bounding box dimensions. It is important to note that this discretization limits the resolution of the components in the generated layouts.

Results of the Docsim measurement between the generated layouts and the conditioning layout can be seen in Table \ref{table:2}. Our model outperforms baseline models in the conditioning scenarios also with respect to this metric. Furthermore, Figure~\ref{fig:res} provides a qualitative comparison of the generated layouts which highlights the common failure cases. For more results, see Appendix A. 

% \begin{table}[!t]
% \centering

% \tablestyle{5.2pt}{1.1}
% \begin{tabular}{lx{22}x{22}x{22}x{22}}
% \shline
% \multicolumn{1}{c}{\textbf{Setting}} & \multicolumn{4}{c}{Publaynet Category+Size}  
%  \\ Model
% & pIOU & Overlap & Alignment & FID  \\
% \shline
% Edit only in inference & 3.3 & 11.5 & 0.24 & 20.1\\
% Without mask embedding & 0.9 & 5.3 & 0.16 & 5.91\\
% DLT & \textbf{0.82} & \textbf{4.6} & \textbf{0.13} & \textbf{5.63} \\

% \end{tabular}

% \caption{Quantitative comparison of different conditioning approaches on the Publaynet dataset in the category+size condition setting. The values of Alignment, Overlap and pIOU are multiplied by 100 for visibility.}
% \label{table:3}
% \end{table}

\begin{table}[!t]
\centering

\tablestyle{5.2pt}{1.1}
\begin{tabular}{lx{22}x{22}x{22}x{22}}
\shline
\multicolumn{1}{c}{\textbf{Setting}} & \multicolumn{4}{c}{Publaynet Category+Size}  
 \\ Model
& pIOU & Overlap & Alignment & FID  \\
\shline
Edit only in inference & 5.8 & 16.6 & 0.19 & 14.1\\
Without condition embedding & 1.1 & 5.3 & 0.17 & 4.3\\
DLT & \textbf{0.7} & \textbf{4.5} & \textbf{0.13} & \textbf{2.9} \\

\end{tabular}

\caption{Quantitative comparison of different conditioning approaches on the Publaynet dataset. Conditioning is done on all the component sizes and categories + half of the component locations. The values of Alignment, Overlap, and pIOU are multiplied by 100 for clarity.}
\label{table:3}
\end{table}

\subsection{Key components Analysis}
In order to investigate the effectiveness of our proposed method, we conduct an ablation study on the key components of the model.

\textbf{Conditioning Mechanism.} We compare our conditioning mechanism, given in Section~\ref{sec:conditioning}, to two alternatives: (1) Performing editing only during inference, and (2) Removing the condition embedding. In (1), we train the model only according to the unconditioned setting (see Section~\ref{4.5}). At the inference step, we override the noise in the reverse diffusion process with the conditioned information. It is important to note that this approach (at the discrete case) is equivalent to the conditioning approach in~\cite{BLT}. In option (2), we still sample during training from the three conditioning settings as described in Section~\ref{4.5}, with a slight modification that we condition also on half of the components (chosen randomly). In this option, we remove the condition embedding. This means that the model does not have explicit information if the input is part of the diffusion process or the real conditioning input. The DLT model was also trained with the same sampling as in option (2).

A comparison of the methods on the Publaynet dataset, conditioning on the Category+Size of all the components and on all the information of half of the components, is given in Table \ref{table:3}. As seen, performing conditioning only during inference significantly degraded the model's performance, whereas adding the condition embedding further improves the results. This information (which component is part of the conditioning) allows the model to resolve the conditioning ambiguity and align the generated components according to the conditioned components. For more details, see Appendix B.

\textbf{Joint Continues-Discrete diffusion process.} In order to evaluate the effectiveness of our proposed joint continues-discrete diffusion process, we compare it to running two sequential diffusion processes independently. In the first compared model, we first run a discrete diffusion process on the category. During training, we replace the unconditioned setting described in Section~\ref{4.5} by conditioning on the box location and size where we set the location and size of all boxes to be zero. In this way, the model is trained on each sample either to reverse the discrete process or the continuous process (but not on both of them). At inference, we first run the discrete diffusion process on the component categories (with the location and size equal to zero), and then given the predicted categories we run the continuous reverse diffusion process conditioning on the reverse discrete process results. In the second compared method, we first run the continuous reverse diffusion process and then run the discrete reverse process. In training, this is done by replacing the unconditioned setting with two settings: (1) Running the continuous diffusion process on the component location + size conditioned on all categories equal to the mask token. (2) Running the discrete diffusion process conditioned on the components' locations + sizes. Observe that also in this method for each sample we train the model to reverse either the discrete or the continuous diffusion process. At inference, we first run the reverse continuous process conditioned on all categories equal to the mask token to predict the component location. Then we run the reverse discrete process to predict the component category conditioned on the box location. 

We provide results for the unconditioned setting over the Publaynet dataset  in Table~\ref{table:4}. As can be seen, our proposed joint continuous-discrete diffusion process outperforms the sequential diffusion process. It is important to note also that the inference time of the joint model is twice faster. 

\begin{table}[!t]
\centering

\tablestyle{5.2pt}{1.1}
\begin{tabular}{lx{22}x{22}x{22}x{22}}
\shline
\multicolumn{1}{c}{\textbf{Setting}} & \multicolumn{4}{c}{Publaynet Unconditioned}  
 \\ Model
& pIOU & Overlap & Alignment & FID  \\
\shline
Class before boxes & 0.59 & 2.1 & 0.11 & 50.8\\
Class after boxes & 0.53 & 2.3 & 0.14 & 20.5\\
DLT (Joint process)& \textbf{0.23} & \textbf{1.8} & \textbf{0.11} & \textbf{20.1}\\ \\

\end{tabular}

\caption{Quantitative comparison of different approaches to combine the discrete and the continuous diffusion processes on the Publaynet dataset in the unconditioned  setting. The values of Alignment, Overlap, and pIOU are multiplied by 100 for clarity.}
\label{table:4}
\end{table}

\section{Conclusions and future work}
In this work, we present a new method for generating synthetic layouts given partial information on the layout structure. We introduce the DLT model, a Diffusion layout transformer that performs a joint continuous-discrete diffusion process on the dimensions and categories of layout components. The model offers flexible conditioning on the input by using a condition embedding that explicitly specifies on which part of the components the diffusion process is applied.  
Experimental results on three public datasets over three different conditioning settings demonstrate the effectiveness and flexibility of our proposed method, and that it outperforms existing methods. %Our approach significantly outperforms existing methods in all the evaluated settings. 
We further investigate the effectiveness of our model's key components. By comparing different alternatives, we show the added value both of the joint continuous-discrete process and the conditioning mechanism.
A limitation of our current work is that the generation process is done without considering the content of the components. This information is very important for resolving many ambiguities in the generation process and can lead to significant improvements in the results. Since the embedding inputs for the DLT transformer backbone represent a layout component, it is natural to extend our solution by encoding more information about the component content in the component embedding.  We intend to further explore this direction in future work.
{\small
\bibliographystyle{ieee}
\bibliography{egbib_final}
}
\pagebreak
$ $

\begin{table*}[!h]
\centering

\tablestyle{5.2pt}{1.25}
\begin{tabular}{lx{22}x{22}x{22}x{22}|x{22}x{22}x{22}x{22}|x{22}x{22}x{22}x{22}}
\shline
\multicolumn{1}{c}{Dataset} & \multicolumn{12}{c}{ \textbf{Publaynet}} \\\cline{2-13}

\multicolumn{1}{c}{} & \multicolumn{4}{c}{Conditioned on Category} & \multicolumn{4}{c}{Category + Size}& \multicolumn{4}{c}{Uncoditioned} \\ Model
& pIOU & Overlap & Alignment & FID & pIOU & Overlap & Alignment & FID &  pIOU & Overlap & Alignment & FID \\
\shline 
LT~\cite{LT} & 0.2 & 0.3 & 0.06 & 1.2 & 0.3 & 0.4 & 0.01 & 1.4 & 0.02 & 0.2 & 0.01 & 1.5\\
BLT~\cite{BLT} & 0.1 & 0.7 & 0.03 & 2.3 & 0.3 & 0.7 & 0.01 & 0.6 & 0.01 & 0.1 & 0.01 & 2.7\\
VTN~\cite{VTN}& 0.2 & 0.5 & 0.03 & 1.1 & 0.4 & 0.4 & 0.01 & 1.0 & 0.01 & 0.1  & 0.01 & 1.9\\
\cline{1-13}
DLT & 0.08 & 0.2 & 0.02 & 0.9 & 0.09 & 0.3 & 0.01 & 0.8 & 0.01 & 0.1  & 0.01 & 0.9\\
\shline

\multicolumn{1}{c}{Dataset} & \multicolumn{12}{c}{ \textbf{Rico}} \\\cline{2-13}

\multicolumn{1}{c}{} & \multicolumn{4}{c}{Conditioned on Category} & \multicolumn{4}{c}{Category + Size}& \multicolumn{4}{c}{Uncoditioned} \\ Model
& pIOU & Overlap & Alignment & FID & pIOU & Overlap & Alignment & FID &  pIOU & Overlap & Alignment & FID \\
\shline
LT~\cite{LT} &  0.9 & 1.6 & 0.02 & 0.5 & 1.0 & 1.1 & 0.06 & 0.3 & 1.5 & 0.7 & 0.01& 0.4\\
BLT~\cite{BLT} & 1.2 & 1.6 & 0.01 & 0.7 & 0.9 & 0.9 & 0.08 & 0.4 & 1.8 & 0.8 & 0.01& 0.9\\
VTN~\cite{VTN}& 1.0 & 1.4 & 0.02 & 0.4 & 1.1 & 0.9 & 0.05 & 0.3 & 1.9 & 0.7 &0.01 & 1.0\\
\cline{1-13}
DLT  &  1.0 & 1.3 & 0.01 & 0.5 & 0.8 & 0.6 & 0.07 & 0.3 & 1.6 & 0.9 & 0.01 & 0.6\\
\shline

\multicolumn{1}{c}{Dataset} & \multicolumn{12}{c}{ \textbf{Magazine}} \\\cline{2-13}

\multicolumn{1}{c}{} & \multicolumn{4}{c}{Conditioned on Category} & \multicolumn{4}{c}{Category + Size}& \multicolumn{4}{c}{Uncoditioned} \\ Model
& pIOU & Overlap & Alignment & FID & pIOU & Overlap & Alignment & FID &  pIOU & Overlap & Alignment & FID \\
\shline
LT~\cite{LT} & 0.8 & 2.9 & 0.4 & 3.1 & 1.1  & 2.4 & 0.3 & 2.7 & 1.8 & 1.6 & 0.3 & 2.7\\
BLT~\cite{BLT} & 1.1 & 3.1 & 0.3 & 2.7 & 1.7 & 4.6 & 0.4 & 2.0 & 2.3 & 2.6 & 0.3 & 3.6\\
VTN~\cite{VTN} & 0.7 & 2.7 & 0.3 & 2.6 & 1.0 & 2.1 & 0.3 & 1.9 &  1.6 & 0.9 & 0.5  & 4.4\\
\cline{1-13}
DLT  & 0.2 & 1.7 & 0.2 & 2.2 & 0.7 & 1.3 & 0.2 & 1.6 & 0.4 & 0.5 & 0.4 & 2.6\\
\shline

\end{tabular}

\vspace{+10pt}
\caption{Std results of all tested experiments in Table~\ref{table:1}}
\label{table:5}
\end{table*}

\pagebreak

\begin{table}[h]
\centering

\tablestyle{5.2pt}{1.1}
\begin{tabular}{lx{22}x{22}x{22}x{22}}
\shline
\multicolumn{1}{c}{\textbf{Setting}} & \multicolumn{4}{c}{Publaynet Category+Size}  
 \\ Model
& pIOU & Overlap & Alignment & FID  \\
\shline
Edit only in inference & 3.3 & 11.5 & 0.24 & 20.1\\
Without condition embedding & 0.86 & 4.8 & 0.15 & 5.55\\
DLT & 0.85 & 5.3 & 0.14 & 5.73 \\

\end{tabular}

\caption{Quantitative comparison of different conditioning approaches on the Publaynet dataset in the category+size condition setting. The values of Alignment, Overlap, and pIOU are multiplied by 100 for clarity.}
\label{table:appendix_cond}
\end{table}

\section*{Appendix}

\subsection*{A. Additional results}
\begin{figure*}[t!]
	\begin{center}
        \includegraphics[width=0.8\linewidth]{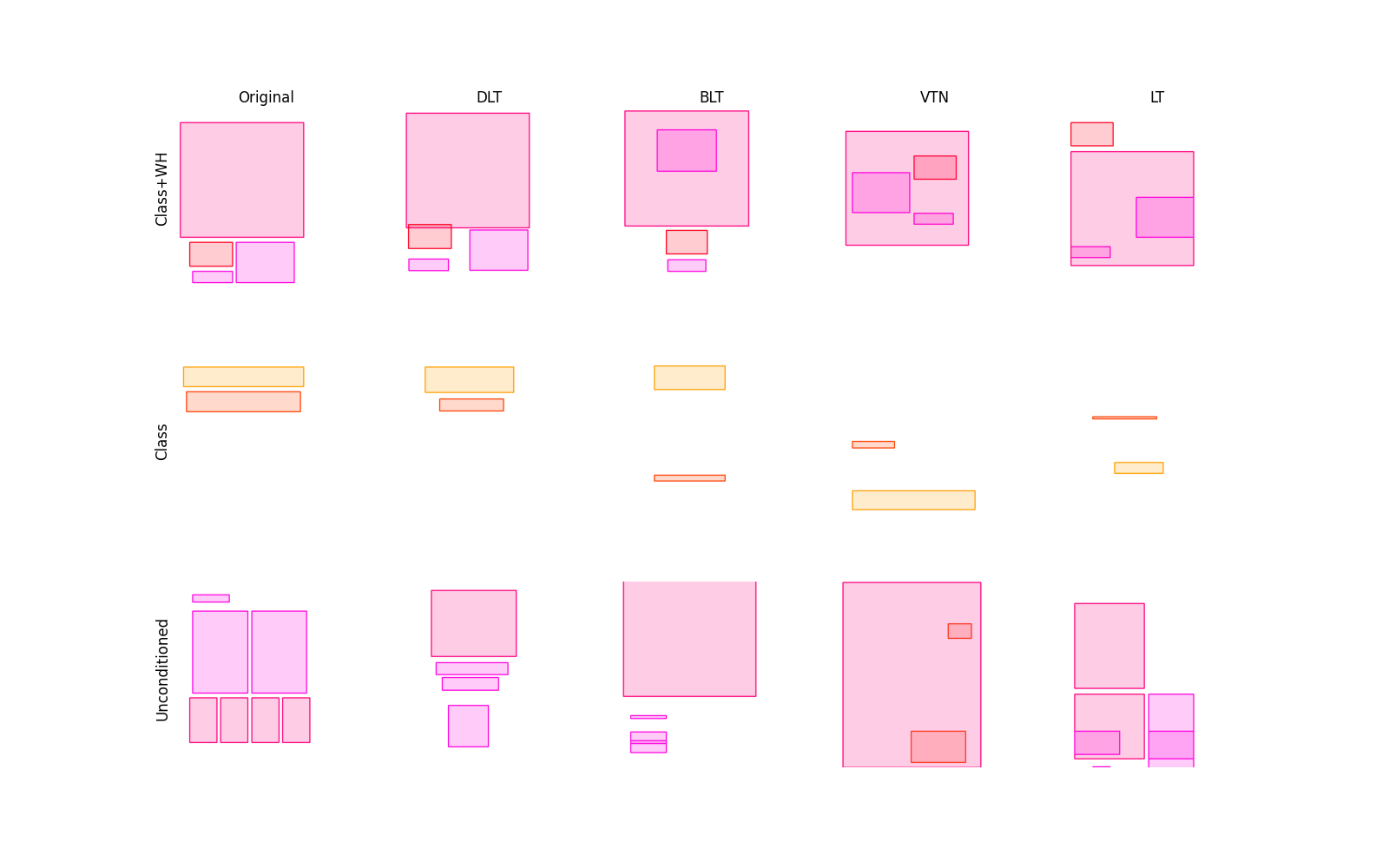}
    %\vspace{-10pt}
	\caption{\small{Qualitative comparison of generated layouts between all tested methods on the Magazine dataset with the three tested conditioning settings. 
	}}
	\vspace{-10pt}
	\label{fig:magazine_res}
	\end{center}
\end{figure*}

We provide additional experimental results. In Table~\ref{table:5} we provide the std of all the experiments in Table~\ref{table:1}. Each of the experiments was tested with four trials on the test data of the relevant dataset. As can be seen, our model outperforms the tested methods by a statically significant margin. 
We also provide an additional qualitative comparison on the Magazine dataset in Figure~\ref{fig:magazine_res}. It is important to note that this dataset is a small dataset which results in degraded performances (compared to a bigger dataset such as PubLaynet). However, our model still performs better also on the Magazine dataset compared to the tested model.

\subsection*{B. Conditioning mechanism experiments}
\begin{figure*}[t!]
	\begin{center}
        \includegraphics[width=0.8\linewidth]{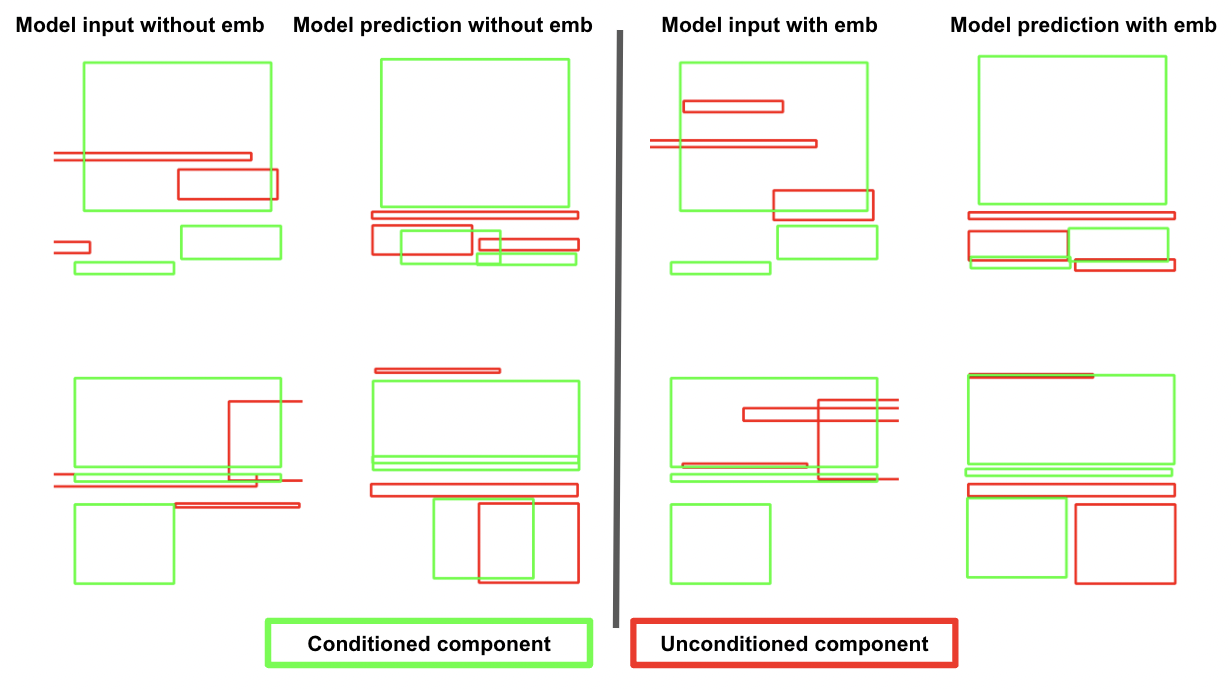}
    %\vspace{-10pt}
	\caption{\small{A comparison of model prediction with and without the condition embedding at the second reverse diffusion step. \textbf{Green:} Components which are part of the conditioning. \textbf{Red:} Components which are part of the generation process. As can be seen, without the embedding the model shifts the conditioned component location and organize the unconditioned part accordingly. With the embedding, the model organizes the unconditioned components according to the conditioned components' locations. 
	}}
	\vspace{-10pt}
	\label{fig:cond_abalation}
	\end{center}
\end{figure*}

In this section, we provide additional information on the conditioning ablation study.
The three initial scenarios described in \ref{sec:conditioning}, were chosen according to~\cite{BLT} evaluation protocol. It is important to note that in this setting each component attribute is either condition on \textbf{all} the components, or predicting the attributes for \textbf{all} the components. In these settings, the condition embedding does not have an impact on the model's performance as can be seen in Table \ref{table:appendix_cond}. However, when the conditioning state of the attributes is not uniform across all components, the condition embedding provides the model with important information that can resolve the alignment ambiguity in the generation process. For example, if we look at the location attributes, as can be seen in Figure \ref{fig:cond_abalation}, without the condition embedding the model shifts the locations of all the components. However, when using the condition embedding the model aligns the unconditioned component according to the condition component. 
 In Table \ref{table:3} we indeed see that in this setting the condition embedding boosts the model performances. 

\end{document}